\documentclass[conference]{IEEEtran}
\IEEEoverridecommandlockouts
\usepackage{cite}
\usepackage{amsmath,amssymb,amsfonts,amsthm}
\newcommand\numberthis{\addtocounter{equation}{1}\tag{\theequation}}
\usepackage{bbm}
\usepackage{algorithm}
\usepackage{algpseudocode}
\usepackage{graphicx}
\usepackage{graphics}
\usepackage{textcomp}
\usepackage{xcolor}
\usepackage{multirow}
\usepackage{colortbl}
\usepackage[normalem]{ulem}
\useunder{\uline}{\ul}{}
\def\BibTeX{{\rm B\kern-.05em{\sc i\kern-.025em b}\kern-.08em
    T\kern-.1667em\lower.7ex\hbox{E}\kern-.125emX}}
\newtheorem{proposition}{Proposition}[section]
\newtheorem{lemma}{Lemma}[section]

\newtheorem{theorem}{Theorem}

\newcommand{\vecX}{\mathbf{x}}
\newcommand{\vecY}{\mathbf{y}}
\newcommand{\vecH}{\mathbf{h}}

\theoremstyle{definition}

\begin{document}

\title{Normalizing self-supervised learning for provably reliable Change Point Detection\\
}


\author{\IEEEauthorblockN{1\textsuperscript{st} Alexandra Bazarova}
\IEEEauthorblockA{\textit{Applied AI Center} \\
\textit{Skoltech}\\
Moscow, Russia \\
A.Bazarova@skoltech.ru}
\and
\IEEEauthorblockN{2\textsuperscript{nd} Evgenia Romanenkova}
\IEEEauthorblockA{\textit{Applied AI Center} \\
\textit{Skoltech}\\
Moscow, Russia \\
shulgina@phystech.edu}
\and
\IEEEauthorblockN{3\textsuperscript{rd} Alexey Zaytsev}
\IEEEauthorblockA{\textit{Applied AI Center} \\
\textit{Skoltech}\\
Moscow, Russia \\
A.Zaytsev@skoltech.ru}
}
\maketitle

\begin{abstract}
Change point detection (CPD) methods aim to identify abrupt shifts in the distribution of input data streams. Accurate estimators for this task are crucial across various real-world scenarios. Yet, traditional unsupervised CPD techniques face significant limitations, often relying on strong assumptions or suffering from low expressive power due to inherent model simplicity.
In contrast, representation learning methods overcome these drawbacks by offering flexibility and the ability to capture the full complexity of the data without imposing restrictive assumptions. 
However, these approaches are still emerging in the CPD field and lack robust theoretical foundations to ensure their reliability.
Our work addresses this gap by integrating the expressive power of representation learning with the groundedness of traditional CPD techniques. We adopt spectral normalization (SN) for deep representation learning in CPD tasks and prove that the embeddings after SN are highly informative for CPD.
Our method significantly outperforms current state-of-the-art methods during the comprehensive evaluation via three standard CPD datasets.


\end{abstract}


\section{Introduction}
Change point detection (CPD) is designed to identify unexpected changes in data streams as quickly and accurately as possible. 
Such a problem statement has been vital for decades for various real-world problems, spanning detailed theoretical analysis and the development of diverse applied methods. The examples include industrial quality control~\cite{industry}, medicine~\cite{med}, video surveillance~\cite{indid}, finance~\cite{finance}, climate change~\cite{climate}, human activity recognition~\cite{activity}, and more.
Most of these problems fall under the unsupervised setting due to the frequent unavailability of annotated data~\cite{ts2vec,indid}. This common scenario necessitates robust techniques that can operate effectively without labeled examples.

Traditionally, CPD problems in the unsupervised setting are solved via different theoretically-justified parametric and non-parametric approaches~\cite{Aminikhanghahi2017-jw, truong2020selective}. 
Although these methods have a solid theoretical base, they suffer from strong assumptions about the underlying distribution and type of changes \cite{enikeeva, gustafsson}.
For example, they look for the change in the mean or the variance in the underlying distribution~\cite{gharghabi, liu2013}.

Such limitations can be handled via deep learning methods: with no assumptions for input data distribution, a well-designed neural network captures the key aspects of time series structure, achieving better detection~\cite{kl-cpd, tscp}. Moreover, they provide the only way to work with complex high-dimensional streams~\cite{indid}.
For the unsupervised setting, self-supervised learning (SSL) methods provide a consistent approach for, e.g., anomaly detection, among other problems. These methods have been widely used and led to advances for specific problems and modalities \cite{byol, ts2vec}.
However, only a few papers on this topic have been published in the context of change point detection~\cite{tscp, vae-cp, cocpd}.
The problem is the complexity of the representation space in an unsupervised regime and no explanation of their behavior even for anomaly detection~\cite{sngp,vazhentsev2022uncertainty}.
Thus, there is no theoretically rigorous understanding of how to construct the most suitable embeddings for successful CPD.

We take the best of both worlds: our method improves state-of-the-art representation learning models and aligns them with classic theory from the CPD area to produce reasonable embedding space. 
Within our approach, the SN forces the representation learning model to keep distributional shifts in embedding space, leading to superior empirical performance. 
The results are supported by theoretical justification and various experiments for different datasets and models. 

\begin{figure}[h]
\centerline{\includegraphics[scale=0.035]{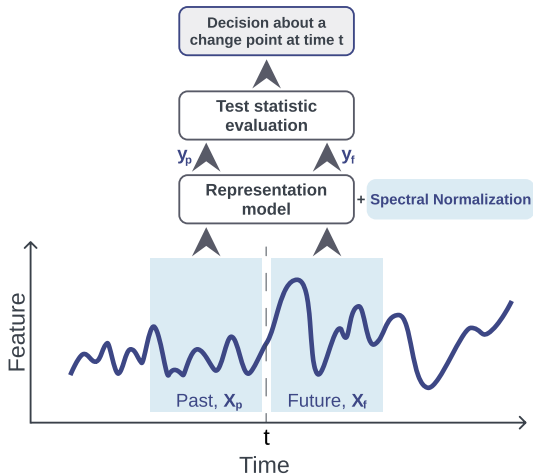}}
\caption{The structure of the proposed change point detection procedure based on a model obtained via self-supervised learning with spectral normalization.}
\label{fig:pipeline}
\end{figure}

Our main contributions are:
\begin{itemize}
    \item A proof that the usage of Spectral Normalization (SN) for neural networks ensures that changes in the raw data are preserved in the representation space, maintaining the test power for standard change point detection methods.
    \item A framework based on self-supervised representation learning augmented with SN for change point detection. As base SSL methods, we consider two models: contrastive TS2Vec \cite{ts2vec} and non-contrastive TS-BYOL~---~our adaptation of BYOL\cite{byol} for time series~data. The pipeline is presented in Figure~\ref{fig:pipeline}.
    \item Empirical evidence on the effectiveness of the proposed framework for three standard CPD datasets. 
\end{itemize}

\section{Related works}
\label{sec:rw}
We focus on the development of CPD methods in the unsupervised setting. 
It is of particular interest since it is often challenging to obtain annotated data due to the complexity and high cost of the annotation process. 
For example, in the oil\&gas domain, well-log data segmentation \cite{wells} requires experts' involvement for annotation. 
Another example is the medical imaging domain \cite{Malladi2013}, where the annotation process is very time-consuming and sometimes even includes the consultation of several experts. Thus, all methods that do not need labels from the experts are valuable from a practical point of view.

A substantial part of the unsupervised CPD methods compares a predefined statistic of time series subintervals with some threshold: if the value exceeds the threshold --- the model alarms a change point. There are different ways to define such statistics, including likelihood ratio as in CUSUM methods~\cite{cusum, cusum1}, the observability matrices-based dissimilarity as in subspaces analysis approaches~\cite{si, sst}, Information entropy~\cite{espresso}, or simple distance in kernel spaces~\cite{Harchaoui2009}. However, all methods described above have significant drawbacks: they all either strongly rely on assumptions about the distribution of the time series data or the state space model, only take into consideration a few characteristics (like temporal shape) of the data, not being able to reflect the complexity of its structure, or are limited by the simplicity of the structure of the selected statistics.

Deep learning-based methods for the CPD task are particularly interesting because they do not require strong assumptions about the input data, are more flexible in terms of architecture than kernel- or graph-based methods, and allow to work with various data types, including images and video~\cite{indid}. 
One example of such a method is KL-CPD \cite{kl-cpd}, which extends kernel-based approaches by considering deep kernel learning; the authors of \cite{kl-cpd} have shown that this method outperforms multiple classical approaches. However, KL-CPD is still bounded by the shape of the kernel and does not fit the unsupervised paradigm. Given the chosen setting, we turn to self-supervised representation learning methods. These methods have shown outstanding performance \cite{ts2vec, byol}, being very versatile and thus suitable for various tasks. In the context of CPD, they are not very widely researched yet. One of the key works in this field is TS-CP2 \cite{tscp}, which first applied contrastive learning to the considered task. Briefly, contrastive learning aims to learn such an embedding space in which similar sample pairs stay close to each other while dissimilar ones are far apart. The authors of \cite{tscp} have shown that this approach is beneficial for change point detection; however, their method, as well as other existing ones in this area, relies on the intuitive explanation of its behavior, lacking a rigorous theoretical foundation. 

Our work takes the next step in developing these methods. We propose applying the spectral normalization (SN) method, which was initially used to stabilize GANs \cite{specnorm}, to the representation learning models. The SN method has already been used as a mechanism for ensuring the invertibility of neural networks \cite{iresnet}; we, in turn, extend this idea to obtain specific statistical properties of the learned representations. This technique allows us to align the expressive power of deep self-supervised learning with the existing mathematical methods, providing theoretical justification for the reasonableness of the learned representations from the change point detection point of view.

\section{Methodology}\label{sec:methodology}
Let us begin with the formal statement of change point detection problem \cite{tartakovsky}. 
As input, we have a sequence of independent $D$-dimensional observations $\vecX_1, \vecX_2, \dots$ that possibly has a change in distribution at an unknown moment $\nu$.
So, observations $\vecX_1, \vecX_2, \dots, \vecX_{\nu}$ come from one distribution, while $\vecX_{\nu + 1}, \vecX_{\nu + 2}, \dots$~---~from another distribution. 
Without loss of generality we can assume that $\vecX_i$ belongs to the $D$-dimensional sphere $
\mathbb{S}^{D - 1}$.
The moment $\nu$ of the abrupt change is called the \textit{change point}, with 
$\nu = \infty$ denoting the absence of a change. The goal is to detect the change point as precisely as possible, avoiding false alarms and long delays. 

Following the state-of-the-art CPD methods \cite{tscp}, we propose to solve this problem in the representations space instead of the raw data space. 
The goal of the transition into the representation space is to capture the underlying structure of the time series, discard measurement artifacts and noise, and reduce data dimensionality. 
The paper\cite{oord2018representation} shows that InfoNCE loss facilitates this by maximizing mutual information between the representations. 

While getting rid of the noise, we need to make sure that nothing important is lost~---~we hope that the representations, if not more informative than the raw data itself, are at least \textit{as informative as the raw data} in the context of the change point detection task. 

To ensure this, we propose applying the spectral normalization (SN) technique \cite{specnorm} on the weight matrices $\{ \mathbf{W}_l\}_{l=1}^{L}$ for an $L$-layered representation learning network. 
At every training step (before the optimization step), the SN method evaluates the spectral norm of the matrix $\hat{\lambda} \approx \lVert \mathbf{W}_l \rVert_2$ via the power iteration method \cite{Mises1929} and then enforces it to be less than a predefined hyperparameter~$c$:
\begin{equation}\label{sn}
\mathbf{W}_l = \begin{cases}
      c \cdot \frac{\mathbf{W}_l}{\hat{\lambda}}, \; \; \text {if $c < \hat{\lambda}$,} \\
      \mathbf{W}_l, \; \; \text{otherwise.}\\
    \end{cases} 
\end{equation}

This technique should be beneficial for representation learning in the context of CPD, as it enhances the test power-preserving property in neural networks. 
Below, we (1) describe the theoretical justification for this argument and (2) provide experimental evidence suggesting that SN benefits the change point detection quality.

\subsection{Spectral Normalization preserves the CPD quality}

Consider the following problem setting. 
Let $\vecX_1, \dots, \vecX_t$ be a sequence of independent observations. 
Before the CP, observations come from the distribution $p_{\infty}$, after --- from $p_{0}$. Both distributions and the change point $\nu$ are considered unknown.
The change point detection problem is a problem of testing a hypothesis $H_0$ versus an alternative $H_1$:
\begin{align*}
    & H_0: \nu = t, \; &&\text{i.e.}\; \vecX_i \sim p_{\infty} \; \; \forall \, 1 \leq i \leq t, & \numberthis \label{eq1}  \\
    & H_1: \nu \in \{1, \dots, t - 1 \}, \; &&\text{i.e.}\; \vecX_{\leq \nu} \sim p_{\infty}, \; \vecX_{>\nu} \sim p_0. & 
\end{align*}

We will consider the two most common approaches in this setting: 
the two-sample tests built on kernel-based statistics and the likelihood ratio-based tests. Below, it is shown that spectral normalization ensures the transition into the space of representations: 
\begin{enumerate}
    \item in the case of two-sample kernel-based tests~---~does not change the type II error rate of convergence to zero (for \underline{no} assumptions about the data distribution at all);
    \item in the case of likelihood ratio-based CPD tests~---~does not decrease the power of such tests (for reasonable assumptions about the data distribution).
\end{enumerate}


But first, let us outline the main property that the spectral normalization equips the neural networks with.




\subsubsection{Bi-Lipschitz neural networks}
The paper \cite{sngp} showed that the SN technique ensures the bi-Lipschitz property for the neural networks $G$ of the form: 
\begin{equation}\label{func}
    G(\mathbf{X}) = h \circ g(\mathbf{X}),
\end{equation}
where $g(\mathbf{X}) = \mathbf{A}\mathbf{X} + \mathbf{B}$, and $h(\mathbf{X})$ is a composition of $L$ residual blocks:
\begin{align}
\label{resblock}
    &h(\mathbf{X}) = h_L \circ \dots \circ h_1(\mathbf{X}), \\ &h_l(\mathbf{X}) = \mathbf{X} + g_l(\mathbf{X}), \; l=1, \ldots, L, \nonumber
\end{align}
where $g_l(\mathbf{X}) = \sigma(\mathbf{W}_l \mathbf{X} + \mathbf{B})$. 
Here, we can consider the case of consecutive mappings of $\vecX_i$ s.t. $\vecY_i = G(\mathbf{X}_{i-w:i})$ and $\mathbf{Y} = \{\vecY_i\}_{i = 1}^t$ or the case when $\mathbf{Y} = G(\mathbf{X})$ is a single vector; i.e., the mapping $G$ may (depending on the specific CPD procedure) have an image in either $\mathbb{R}^d$ or $\mathbb{R}^{t \times d}$.
Inside $g_l(\mathbf{X})$, $\sigma$ is a sigmoid function, but other non-linearity functions are possible.
Neural networks of such form are common among the modern methods~\cite{vaswani, he2016deep}. 


More precisely, the authors of \cite{sngp} proved the following lemma: 
\begin{lemma}\label{lem:bi-lipsch} \cite{sngp}
    Consider a hidden mapping $h: \mathcal{X} \rightarrow \mathcal{H}$ of a form \eqref{resblock}. If for $0 < \alpha \leq 1$ all $g_l$'s are $\alpha$-Lipschitz, i.e., $\lVert g_l(\mathbf{X}) - g_l(\mathbf{X}') \rVert_{\mathcal{H}} \leq \alpha \lVert \mathbf{X} - \mathbf{X}' \rVert_{\mathcal{X}} \; \forall (\mathbf{X}, \, \mathbf{X}') \in \mathcal{X}^2$. Then:
    \begin{equation}
        L_1 \lVert \mathbf{X} - \mathbf{X}' \rVert_{\mathcal{X}} \leq \lVert h(\mathbf{X}) - h(\mathbf{X}') \rVert_{\mathcal{H}} \leq L_2 \lVert \mathbf{X} - \mathbf{X}' \rVert_{\mathcal{X}}, 
    \end{equation}
    where $L_1 = (1 - \alpha)^L, \, L_2 = (1 + \alpha)^L$.
\end{lemma}

Let $\mathcal{X} = \mathcal{H} = \mathbb{R}^n,$ and $\lVert . \rVert_{2}$ denote the Euclidean distance. Since the SN technique enforces $\lVert \mathbf{W}_l \rVert_2 \leq c$, and the Lipschitz constant of the residual block \eqref{resblock} is bounded by~$\lVert \mathbf{W}_l \rVert_2$:
\[
    \lVert g_l(\mathbf{X}) - g_l(\mathbf{X}') \rVert_2 \leq \lVert \mathbf{W}_l \mathbf{X} - \mathbf{W}_l \mathbf{X}' \rVert_2 \leq \lVert \mathbf{W}_l \rVert_2 \lVert \mathbf{X} - \mathbf{X}' \rVert_2.
\]
With $c \leq 1$, the spectral normalized hidden mapping $h$ ensures that the Euclidean norm is preserved; the same is true for the aforementioned mapping $G(\mathbf{X}) = h \circ g(\mathbf{X})$. Note that in finite dimensional vector spaces, all norms are equivalent; hence, the preservation of the Euclidean distance implies the preservation of any vector norm distance. We will say that the mappings for which lemma \ref{lem:bi-lipsch} holds are \textit{distance-preserving}.

\subsubsection{Preservation of kernel distances}
In change point detection, the test statistic $S(\mathbf{X})$ is often a function of kernel distances between the observations $k(\vecX, \vecX')$ \cite{Harchaoui2009, mmd}. 
Most popular kernels, such as RBF $k(\vecX, \vecX') = \exp \left(\frac{-\lVert \vecX - \vecX'\rVert^2_2}{2\sigma^2} \right)$, rely on the vector norm of the difference $\lVert \vecX - \vecX' \rVert$. In the previous section, we proved the bi-Lipschitz property of the SN networks, i.e., the preservation of such distances. Then, the following statement holds.
\begin{lemma}\label{lem:rbf_distance}
    Consider two observations $\vecX_i, \, \vecX_j \in \mathbb{S}^{D-1}$ and their representations $\vecY_i, \, \vecY_j$.  Consider the RBF kernel $k(\vecX, \vecX')$. SN networks preserve RBF kernel distance, i.e. $\exists \hat{C}, \tilde{C} > 0$: $$\hat{C} k(\vecX_i, \vecX_j) \leq k(\vecY_i, \vecY_j) \leq \tilde{C} k(\vecX_i, \vecX_j).$$
\end{lemma}
 We provide the proof in the Appendix \ref{app:rbf}. Note that this statement holds for both $G: \mathcal{X} \rightarrow \mathbb{R}^d$ and $G: \mathcal{X} \rightarrow \mathbb{R}^{t\times d}$. 
 
\subsubsection{Kernel-based test}
In our experiments, we consider the MMD (\textit{maximum mean discrepancy})\cite{mmd} as a kernel-based statistic. This is a nonparametric probabilistic distance commonly used in two-sample tests. Given a kernel $k$ of the RKHS $\mathcal{H}_k$, the MMD distance between two distributions $P$ and $Q$ is defined as
\begin{align} \label{eq:mmd}
& \mathrm{MMD}^2_k(P, \, Q) = \lVert \mu_P - \mu_Q \rVert^2_{\mathcal{H}^k} = \\ &= \mathbb{E}_P [ k(\zeta, \zeta')] - 2 \mathbb{E}_{P, \, Q}[k(\zeta, \, \xi)] + \mathbb{E}_Q [k(\xi, \xi')],  \nonumber
\end{align}
where $\mu_P = \mathbb{E}_{\zeta \sim P}[k(\zeta, .)], \, \mu_Q = \mathbb{E}_{\xi \sim Q}[k(\xi, .)]$ are the kernel mean embeddings of distributions $P$ and $Q$, accordingly. We use the biased empirical estimate of the $\mathrm{MMD}$ distance: given observations $Z = (\zeta_1, \dots, \zeta_m) \sim P, \, \Xi = (\xi_1, \dots \xi_m) \sim Q$
\begin{align}\label{eq:mmd_approx}
   & \mathrm{MMD}^2_b(Z, \, \Xi) = \frac{1}{m^2}\sum_i\sum_j k(\zeta_i, \zeta_j) - \\ & -\frac{2}{m^2} \sum_i \sum_j k(\zeta_i, \xi_j) + \frac{1}{m^2}\sum_i\sum_j k(\xi_i, \xi_j). \nonumber
\end{align}
For kernels that are preserved under distance-preserving mappings, the following theorem holds.

\begin{theorem}\label{lem:mmd_lemma}
    Consider two sequences of observations $\mathbf{X} = [\vecX_1, \dots, \vecX_n]$ and $\hat{\mathbf{X}} = [\hat{\vecX}_1, \dots, \hat{\vecX}_n], \vecX_i, \hat{\vecX}_i \in \mathbb{S}^{D - 1}$. Denote by $\mathbf{Y}, \, \hat{\mathbf{Y}}$ their images under the distance-preserving mapping~$G$. Consider a bounded kernel $0 \leq k(\vecX, \vecX') \leq K$ that is preserved under $G$ and the corresponding sample $\mathrm{MMD}^k_b$ statistic.
    The $\mathrm{MMD}^{k}_b$-based two-sample test of level $\alpha$ has the same type II error rate $O(n^{-\frac{1}{2}})$ of convergence to zero in the space of embeddings $\mathcal{Y}$ as in the space of raw observations $\mathcal{X}$.
\end{theorem}
For proof, see Appendix \ref{app:mmd_lemma}. Note that the $\mathrm{MMD}$ statistic based on RBF kernels is a particular case of Theorem \ref{lem:mmd_lemma}; we use this statistic in our experiments. 

To sum up, spectral normalization ensures that SN mappings preserve kernel distances, which, in turn, leads to the preservation of $\mathrm{MMD}_b$-based two-sample test properties in the latent space.

\subsubsection{Likelihood ratio tests}

Suppose that $p_0, \, p_{\infty}$ in the setting \eqref{eq1} are known. Consider the likelihood ratio of an interval $\mathbf{X} = [\vecX_1, \dots, \vecX_t] \in \mathbb{R}^{t \times D}$ in the setting \eqref{eq1} as its test statistic:
\begin{equation}
    S(\mathbf{X}) = \frac{p^{[1:t]}_0(\vecX_1, \dots, \vecX_t)}{p^{[1: t]}_{\infty}(\vecX_1, \dots, \vecX_t)}.
\end{equation}

The following proposition gives sufficient conditions for the neural network $G:  \mathbb{R}^{t \times D} \rightarrow \mathbb{R}^{t \times d}$ of the form \eqref{func} to preserve likelihood ratio for elliptical distributions.

\begin{proposition}\label{prop:lr_preserve}
Consider a sequence of independent variables $\mathbf{X} = [\vecX_1,\dots, \vecX_t]$ distributed according to an elliptical distribution with a possible change in mean at an unknown location $\nu$. Consider $G(\mathbf{X})$ of the form \eqref{func}, with $h(\mathbf{X})$ being an invertible function. Denote by $\mathbf{Y} = [\vecY_1, \dots, \vecY_t]$ the transformation of the original sequence, i.e. $\mathbf{Y} = G(\mathbf{X})$.

Let $p_0$ and $p_{\infty}$ be the joint PDFs tested for the original sequence $\mathbf{X}$, $\hat{p}_0$ and $\hat{p}_{\infty}$ be the corresponding joint PDFs for the transformed sequence $\mathbf{Y}$. Then
\begin{equation}
    \frac{p_0(\mathbf{X})}{p_{\infty}(\mathbf{X})} = \frac{\hat{p}_0(\mathbf{Y})}{\hat{p}_{\infty}(\mathbf{Y})}.
\end{equation}
\end{proposition}

The proof is presented in the Appendix. 

Therefore, for the neural network $G(\mathbf{X})$ to preserve likelihood ratio, it is sufficient for its component $h(\mathbf{X})$ to be invertible. Thankfully, the SN technique provides this property. 
As mentioned, the SN enforces the Lipschitz constant of $g_l$ to be not greater than $c$ for any predefined $c > 0$. The authors of \cite{iresnet} show that for $c < 1$, such restraint ensures the invertibility of the residual block $h_l$ \eqref{resblock}. Hence, applying SN to all layers $h_1, \dots, h_L$ enforces the invertibility of $h(X)$. 

This likelihood ratio preservation property entails the following theorem.

\begin{theorem}\label{lem:test_power}
    Consider a statistical CPD test based on the likelihood ratio. The proposed strategy ensures that for any elliptical distribution, the power of such test of level $\alpha$ preserves, i.e.,
    \normalfont
    \begin{align}
    & \mathbb{P}_{H_1}(\textrm{reject } H_0 \, | \,  \text{raw data space}) = \\ = \, & \mathbb{P}_{H_1}(\text{reject } H_0 \,  | \, \text{representation space}). \nonumber
    \end{align}
\end{theorem}
The proof is given in Appendix \ref{app:lemma}.

\textbf{\textit{Summary.}} The SN technique ensures the test power-preserving property in neural networks. This property is valuable since it encourages the results of any considered statistical test of considered form not to deviate too much when carried out in the latent space instead of the observation space. Test power preservation is in some way similar to information preservation but through the lens of the change point detection task.

\subsection{General pipeline}
Now that we have outlined the theoretical properties of the proposed approach let us describe the general CPD pipeline depicted in Figure \ref{fig:pipeline} that we follow. The pipeline consists of three stages: (1) representation construction, (2) test statistic calculation, and (3) change point detection.

In the first stage, following the standard sliding window technique, the raw time series data is cropped into a sequence of overlapping time intervals $\mathbf{X}$ of length $2 w$; for each interval, we consider separately its first $\mathbf{X}_\text{past} \in \mathbb{R}^{w \times D}$ and second part $\mathbf{X}_\text{future} \in \mathbb{R}^{w \times D}$. Then, each $\mathbf{X}_\text{past}$ and $\mathbf{X}_\text{future}$ is transformed into its representation. Denote the embeddings by $(\mathbf{Y}_{\text{past}}, \,  \mathbf{Y}_{\text{future}})$. 

In the second stage, for each pair $(\mathbf{Y}_\text{past}, \, \mathbf{Y}_\text{future})$, the test statistic is calculated. We consider two test statistics: the cosine distance and the MMD-score.

\begin{figure}[h]
\centerline{\includegraphics[scale=0.24]{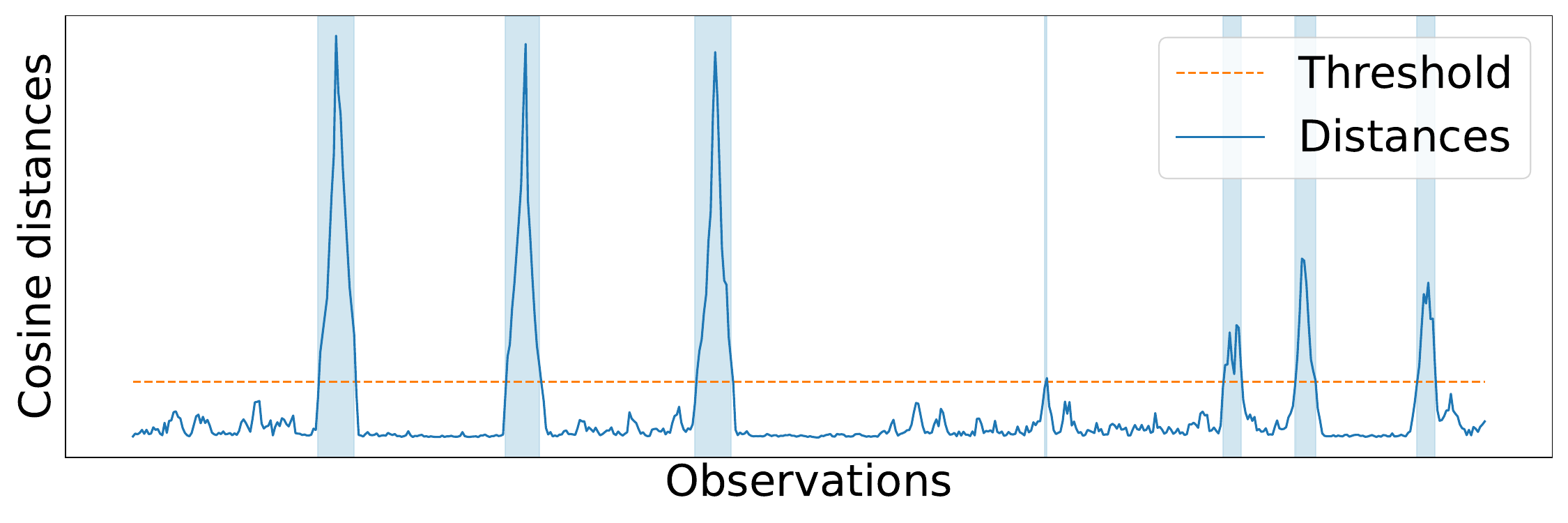}}
\caption{Cosine distances between the subsequent subintervals of observations. The value above a predefined threshold indicates the presence of a changepoint. The color filled area denotes that the corresponding intervals contain a change point.}
\label{fig:similarities}
\end{figure}

In the last stage, the change point detection, we estimate the change points from the obtained statistic values. 
We report a change at every pair of intervals where the statistic is greater than some empirically chosen threshold $\delta$: an example is shown in Figure \ref{fig:similarities}.

\subsection{Models}
As base models, we consider two self-supervised models: TS2Vec~\cite{ts2vec} and our adaptation of BYOL~\cite{byol} for time series data. During the CPD procedure, the embeddings of TS2Vec are obtained in two modes: either an interval is mapped to its embedding, i.e., the model performs the mapping $G: \mathbb{R}^{w \times D} \rightarrow \mathbb{R}^d$~---~and the corresponding test statistic is the cosine distance~---~or an interval is mapped into a sequence of embeddings, i.e., $G: \mathbb{R}^{w \times D} \rightarrow \mathbb{R}^{w \times d}$~---~and the MMD-score is used as the statistic. For BYOL, only the mapping to a single embedding is performed to calculate then the cosine distance. 

\subsubsection{TS2Vec}
\begin{figure}[b!]
\includegraphics[width=\columnwidth]{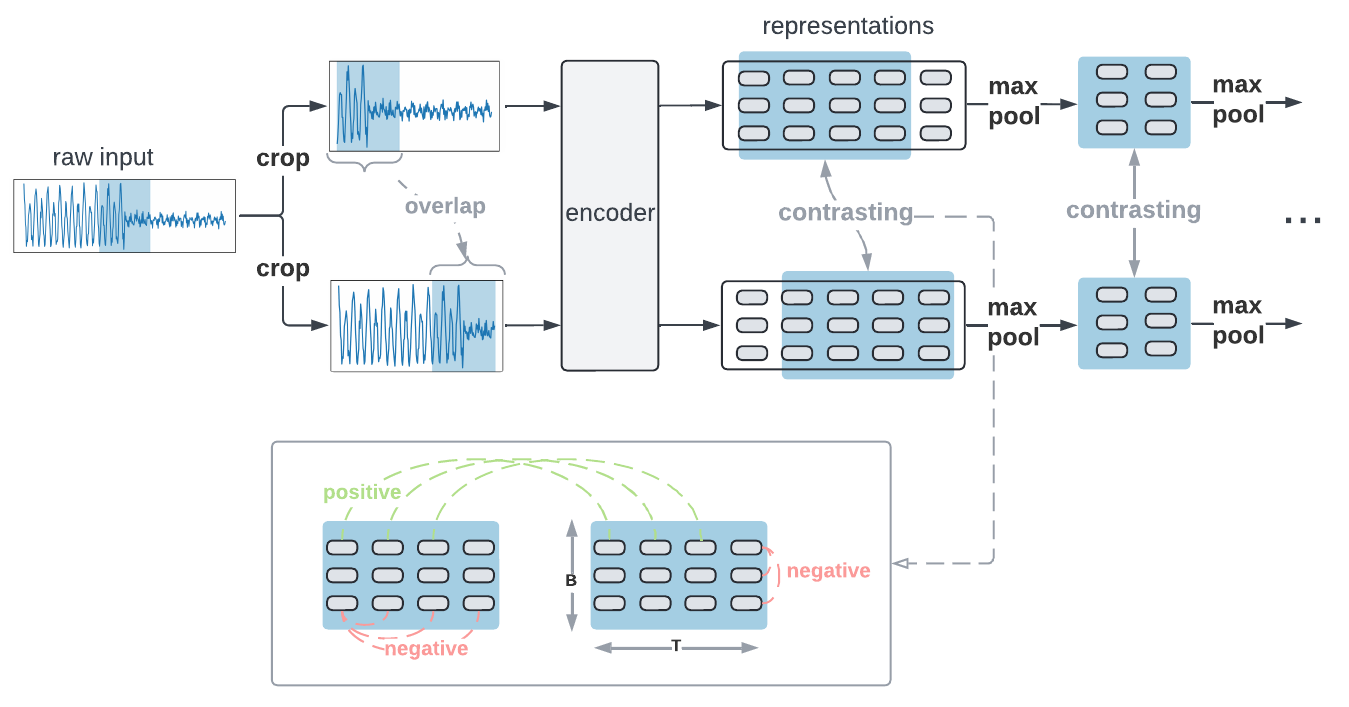}
\caption{TS2Vec architecture. The upper part of the figure is the hierarchical contrasting procedure; the lower part is the selection of positive and negative pairs.}
\label{fig:ts2vec}
\end{figure}

~\cite{ts2vec} leverages hierarchical contrastive loss for learning informative time series representations. The structure of the model is depicted in Figure \ref{fig:ts2vec}. 

In the TS2Vec encoder learning pipeline, the input time series is cropped into two overlapping subseries, creating augmented views of the original data. These samples are then encoded into timestamp-wise sequences of embeddings. The resulting embeddings are contrasted in two dimensions: temporal and instance-wise. The loss for the $t$-th timestamp of the $i$-th sample in a batch includes two corresponding terms:
\begin{align*}
\label{eq:inst_component}
   \mathcal{L}^{(i, t)}_{\text{inst}} &= -\log \frac{\exp (\vecH_{i, t} \cdot \vecH'_{i, t})}{S^{\mathrm{inst}}_i}, \\
   S^{\mathrm{inst}}_i &= \sum_{j = 1}^B \bigl( \exp(\vecH_{i, t} \cdot \vecH'_{j, t}) + \mathbbm{1}(i \neq j)\exp (\vecH_{i, t} \cdot \vecH_{j, t})\bigr); \\ \nonumber
   \mathcal{L}^{(i, t)}_{\text{temp}} &= -\log \frac{\exp (\vecH_{i, t} \cdot \vecH'_{i, t})}{S^{\mathrm{temp}}_i}, \\
   S^{\mathrm{temp}}_i &= \sum_{t' \in \Omega} \bigl( \exp(\vecH_{i, t} \cdot \vecH'_{i, t'}) + \mathbbm{1}(t \neq t')\exp (\vecH_{i, t} \cdot \vecH_{i, t'})\bigr). \nonumber
\end{align*}
Here, $\vecH_{i, t}$ and $\vecH'_{i, t}$ refer to the projections of representations of the same timestamp but from two different augmentations of the original sample; $S^{\mathrm{inst}}_i$  is the sum over the batch; $\Omega$ refers to the overlapping part of the two subseries. 

The contrasting is performed hierarchically: after each iteration of loss calculation, the max pooling operation is executed along the time axis until the dimension of the time axis reduces to one. The final loss comprises of multiple terms corresponding to different levels of representations' granularity.

\subsubsection{TS-BYOL}
\begin{figure}[b!]\label{fig3}
\includegraphics[width=\columnwidth]{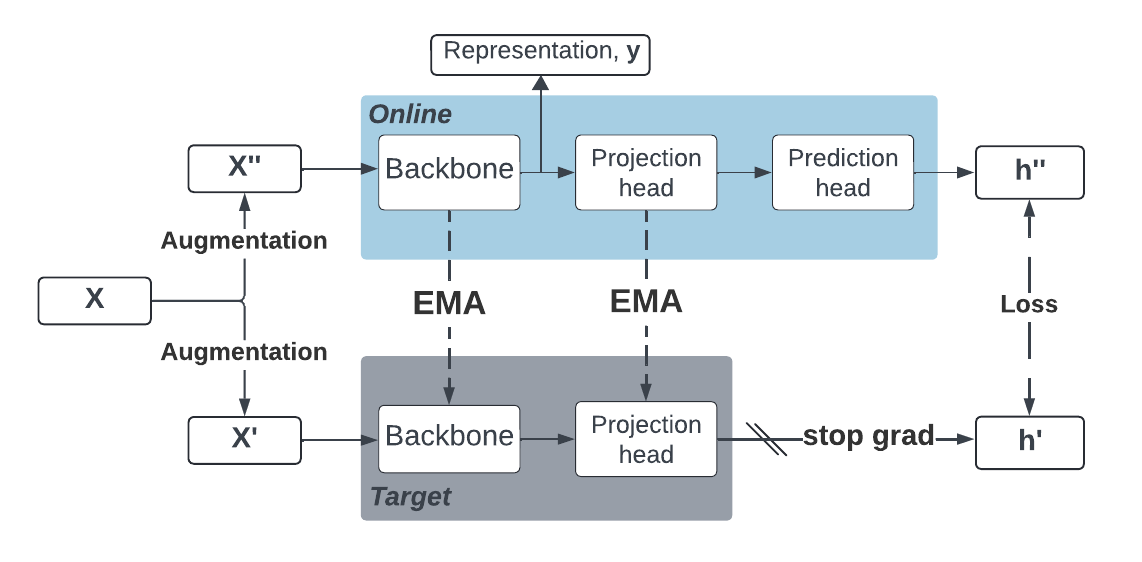}
\caption{BYOL's architecture. The upper part is the training \textit{Online} network, and the lower part is the \textit{Target} network with disabled backpropagation.}
\label{fig:byol}
\end{figure}

We adapt the self-supervised BYOL \cite{byol} model, originally designed for images, as an alternative source of representations. The model (Figure \ref{fig:byol}) consists of two asymmetrical parts: the "target" and "online" networks. The online network has an additional prediction head and is trained via backpropagation, while the target network's weights are updated slowly using exponential moving average (EMA):
\begin{equation}
\label{ema}
    \xi := \beta \xi + (1 - \beta) \theta,
\end{equation}
where $\xi$ and $\theta$ denote the weights of the target and the online network, accordingly, and the hyperparameter $\beta$ is usually close to $1$. 
In our experiments, we set $\beta = 0.996$.

Online and target networks build embeddings $\vecH', \vecH''$ from augmented views $\mathbf{X}', \mathbf{X}''$ of the same sample $\mathbf{X}$. The online network minimizes the $L_2$-norm difference between normalized embeddings $\bar{\vecH} = \frac{\vecH}{\lVert \vecH \rVert_2}$ to match the target network's embeddings:
\begin{equation}
    \mathcal{L}(\vecH', \vecH'') = \lVert \bar{\vecH}' - 
\bar{\vecH}'' \rVert_2^2 = 2 - 2 \frac{\langle \vecH', \vecH'' \rangle}{\lVert \vecH'\rVert_2 \lVert \vecH'' \rVert_2}.
\end{equation} 

For the augmentations, we use only the random cropping of length $w$ from an interval of length $2w$, as a similar procedure provides optimal performance in~\cite{marusov2023noncontrastive}.
The architecture consists of a sequence of 1-D convolutions, followed by ReLU nonlinearities and dropouts, as a backbone; both projection and prediction heads are two-layer MLPs. 

During CPD, we use representations $\vecY$ produced by the backbone neural network, not by the projection heads, and compute cosine distances between the representations of intervals of neighbor windows.

\section{Results}

\subsection{Compared methods} 

\begin{itemize}
    \item \textbf{TS-CP2} \cite{tscp} is a self-supervised model intended for change point detection. It was the first method to leverage contrastive learning to obtain informative time series data representations.
    \item \textbf{KL-CPD} \cite{kl-cpd} is a principled CPD framework that employs deep kernel learning for two-sample hypothesis testing, being another option for NN-based CPD. 
    \item \textbf{ESPRESSO} \cite{espresso} is a hybrid approach that exploits both statistical and temporal shape properties in the CPD process. This method does not involve deep learning while showing decent performance.
    \item A self-supervised \textbf{TS2Vec} \cite{ts2vec} employs hierarchical contrastive loss for representation learning, showing state-of-the-art results in time series classification, forecasting and anomaly detection. \textbf{SN-TS2Vec} is our modification of the TS2Vec model: we performed spectral normalization for each of its convolutional layers. 
    \item An SSL approach \textbf{BYOL}~\cite{byol} provides an additional way to obtain a representation learning model via a non-contrastive loss function. We employ \textbf{SN-BYOL} in some experiments to compare SN-equipped version to a vanilla one. 
\end{itemize}

Suffixes \textit{cos} and \textit{MMD} denote the type of test statistic that is used during the change point detection procedure: either the cosine distance or the MMD score, accordingly. 

\subsection{Datasets}
We provide the analysis of our method on the datasets below. These datasets are considered, following TS-CP2 \cite{tscp}, since we aimed to compare this SOTA model to our method. All datasets are publicly available. 
\begin{itemize}
    \item \textbf{Yahoo} \cite{yahoo}. It is a widely used anomaly detection benchmark consisting of time series that contain metrics of the various Yahoo services with manually labeled anomalies. Following \cite{tscp}, we used the fourth benchmark since it includes annotations of change points.
    \item \textbf{HASC} \cite{hasc}. This dataset contains human activity data collected by three-axis accelerometers. We used the same subset as \cite{kl-cpd} and \cite{tscp}. Change points in this time series indicate alternations in the type of activities (stay, walk, jog, skip, stair up, and stair down).
    \item \textbf{USC-HAD} \cite{usc-had}. The dataset also contains human activity data monitored by wearable sensors. The observed data represents basic daily life activities such as walking, sitting, sleeping, etc. We followed the pipeline from \cite{tscp} and combined 30 random activities from six participants, using only the data from the accelerometer.
\end{itemize}
\vspace{-2mm}
\begin{table}[htbp]
\caption{Datasets}
\begin{center}
\begin{tabular}{lcccc}
\hline
Dataset & Domain& \#Timestamps \footnotemark & \#Sequences  & \#CP \footnotemark\\
\hline
\textbf{Yahoo}& $\mathbb{R}^+$ &  164K & 100 & 208 \\
\textbf{USC-HAD}& $\mathbb{R}^3$ &  97K & 6 & 30 \\
\textbf{HASC}& $\mathbb{R}^3$ &  39K  & 1 & 65 \\
\hline
\end{tabular}
\label{tab1}
\end{center}
\end{table}
\footnotetext[1]{Total number of timestamps.}
\footnotetext[2]{Total number of change points.}

\subsection{$F_1$-score for change point detection}

Since we aim to compare our model to the current SOTA in this field, TS-CP2 \cite{tscp}, we used their equivalent of F1-score for the change point detection task as the primary metric of detection quality
This procedure's inputs are a sequence of binary ground truth labels (for each interval, whether it contains a change point or not) and a sequence of predicted labels, and the output is a value between $0$ and $1$. 
The metric suggests that in order to correctly detect a change point, it is sufficient for the detection algorithm to indicate the presence of a change point in at least one of the consecutive intervals containing it. For extra details, see the TS-CP2 paper \cite{tscp}.

\subsection{Main results} 

Table \ref{table:ts2vec-vs-all} compares SN-TS2Vec to three other state-of-the-art CPD methods. 
For both Yahoo!A4Benchmark and USC-HAD datasets, the SN-TS2Vec approach with the cosine distance used as a test statistic achieves two of the three best results.
For the HASC dataset, all TS2Vec variants show results close to KL-CPD and ESPRESSO, outperforming the TS-CP2 approach. 
So, SN-TS2Vec either dominates the existing methods or performs almost as well as they do.

As for the TS-BYOL model, even though it does not achieve SOTA performance, it shows solid results on the Yahoo!A4Benchmark dataset. Comparison of the SN-BYOL to its vanilla version confirms that the SN technique enhances model properties in the context of CPD task.

\begin{table}[h]
\caption{$F_1$ measure for different detection margins for the proposed SN-TS2VEC approach vs existing CPD methods.}
\centering
\begin{tabular}{ccccc}
\hline
Dataset & Model & \multicolumn{3}{c}{Detection margin} \\ 
\cline{3-5}
& & 24 & 50 & 75 \\ \hline
& TS-CP2 & 0.64 & \textbf{0.81} & 0.843 \\
& KL-CPD & 0.579 & {0.576} & 0.544 \\
& ESPRESSO & 0.224 & 0.340 & 0.4442 \\
& TS-BYOL & 0.5 & 0.706 & 0.768 \\ 
& SN-BYOL & 0.694 & 0.766 & 0.789 \\ 
& TS2Vec & 0.688 & {\ul 0.788} & 0.816\\
& SN-TS2Vec, cos & \textbf{0.726} & 0.775 & \textbf{0.852}\\
{\multirow{-8}{*}{Yahoo}} & SN-TS2Vec, MMD & {\ul 0.692} & \textbf{0.81} & {\ul 0.851} \\
\hline
\multicolumn{2}{l}{}                                              & 100 & 200 & 400 \\ \hline
& TS-CP2 & 0.824 & 0.857 & 0.833 \\
& KL-CPD & 0.743 & 0.718 & 0.632 \\
& ESPRESSO & 0.633 & 0.833 & 0.833\\
& TS-BYOL & 0.5 & 0.796 & 0.933 \\ 
& SN-BYOL & 0.5 & 0.636 & 0.722 \\ 
& TS2Vec & {\ul 0.873} & \textbf{0.97} & {\ul 0.952} \\
& SN-TS2Vec, cos & 0.736 & {\ul 0.909}   & \textbf{1}  \\
{\multirow{-8}{*}{USC-HAD}} & SN-TS2Vec, MMD & \textbf{0.909} & 0.809 & \textbf{1} \\ 
\hline
\multicolumn{2}{l}{}                                              & 60 & 100 & 200 \\ 
\hline
& TS-CP2 & 0.4 & 0.438 & 0.632 \\
& KL-CPD & \textbf{0.479} & \textbf{0.473} & 0.467\\
& ESPRESSO & 0.288 & 0.423 & \textbf{0.693} \\
& TS-BYOL & 0.316 & 0.398 & 0.26 \\ 
& SN-BYOL & 0.403 & 0.416 & 0.418 \\ 
& TS2Vec & {\ul 0.476} & {\ul 0.467} & 0.444 \\
& SN-TS2Vec, cos &  {\ul 0.476} &  0.306 & {\ul 0.663}\\
{\multirow{-8}{*}{HASC}} & SN-TS2Vec, MMD & {\ul 0.476} & {\ul 0.467} & 0.444 \\ 
\hline
\end{tabular}
\label{table:ts2vec-vs-all}
\end{table}

\subsubsection{TS2Vec with and without SN}

The results of the comparison of the model modifications are presented in Table \ref{table:ts2vec-vs-sn-ts2vec}. The experiments were conducted for three different window sizes for each dataset; the hidden representations size was set to $16$ for the Yahoo and USC-HAD datasets and to $32$ for the HASC dataset. On the Yahoo!A4Benchmark dataset, the SN-TS2Vec strongly outperforms its vanilla version, with the cosine distance version being slightly dominant over the MMD-score one. The results on USC-HAD also show the superiority of the SN version: two out of three best results for each dataset belong to it. On the HASC dataset, all models demonstrate comparable results.

\begin{figure}[h]
\centerline{\includegraphics[scale=0.29]{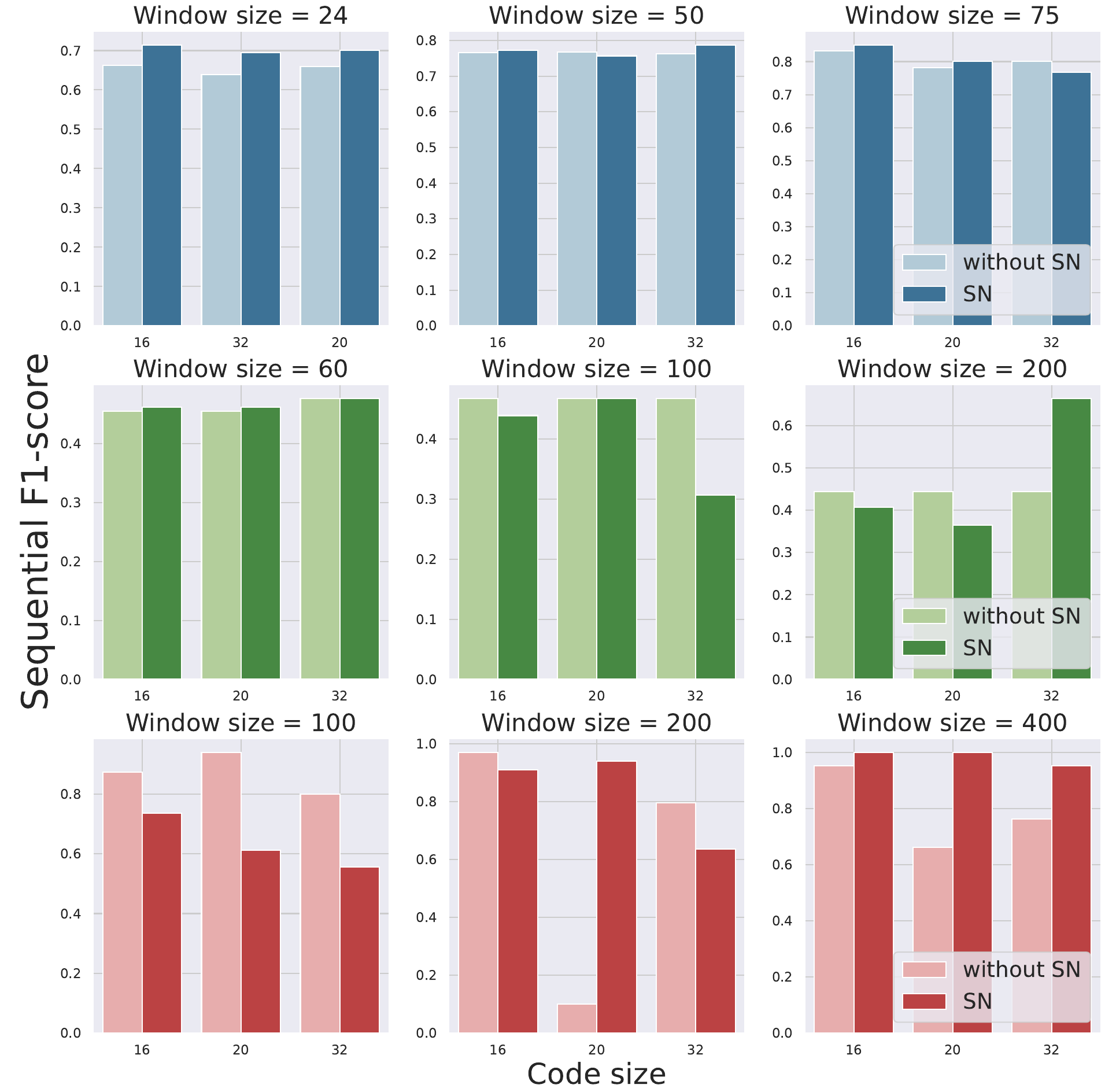}}
\caption{Comparison of TS2Vec performance with and without spectral normalization on all datasets. The cosine distance was used as the test statistic. The upper row~---~Yahoo!A4Benchmark, middle~---~HASC, lower~---~USC-HAD.}
\label{fig:ts2vec_results}
\end{figure}

We also provide the extended version of this comparison in Figure \ref{fig:ts2vec_results}, considering multiple code sizes (i.e., hidden dimensions). The results on the Yahoo!A4Benchmark confirm that the SN technique enhances the properties of the representations; for two other datasets, the SN improves the results for approximately half of the code sizes. Note that these two datasets contain significantly fewer change points than Yahoo!A4Benchmark. Therefore, the results on the latter are more representative.

\begin{table}[h]
\caption{Comparison of vanilla TS2Vec vs SN-TS2vec performance for two test statistics: cosine distance and MMD-score.}
\hspace{2mm}
\begin{tabular}{lccc}
\hline
Model & \multicolumn{3}{c}{Detection margin} \\
\hline
\multicolumn{4}{c}{Yahoo} \\

& 24 & 50 & 75 \\ \cline{1-4}
{TS2Vec, cos}            & {0.688$\pm$0.035}          & {{\ul 0.788$\pm$0.021}}   & 0.816$\pm$0.019           \\
{TS2Vec, MMD}            & {0.624$\pm$0.048}          & {0.764$\pm$0.03}          & 0.796$\pm$0.033           \\
{SN-TS2Vec, cos}         & {\textbf{0.726$\pm$0.021}} & {0.775$\pm$0.018}         & \textbf{0.852$\pm$0.001}  \\
{SN-TS2Vec, MMD}         & {{\ul 0.692$\pm$0.115}}    & {\textbf{0.81$\pm$0.034}} & {\ul 0.851$\pm$0.037}     \\ 
\hline
\multicolumn{4}{c}{USC-HAD} \\                   
& 100 & 200 & 400 \\ \cline{1-4}
{TS2Vec, cos}            & {{\ul 0.873$\pm$0.063}}          & {\textbf{0.97$\pm$0.052}} & {\ul 0.952$\pm$0.083}   \\
{TS2Vec, MMD}            & {0.724$\pm$0.132}          & {{{\ul 0.909$\pm$0}}}       & \textbf{1$\pm$0}         \\
{SN-TS2Vec, cos}         & {0.736$\pm$0.212}          & {{\ul 0.909$\pm$0}  }     & \textbf{1$\pm$0}         \\
{SN-TS2Vec, MMD}         & {\textbf{0.909$\pm$0}}        & {0.809$\pm$0}             & \textbf{1$\pm$0}        \\ 
\hline
\multicolumn{4}{c}{HASC} \\                       & 60 & 100 & 200 \\ 
\hline
{TS2Vec, cos}            & {\textbf{0.476$\pm$0}}     & {\textbf{0.467$\pm$0}}             & {\ul 0.444$\pm$0}        \\
{TS2Vec, MMD}            & {\textbf{0.476$\pm$0}}     & {\textbf{0.467$\pm$0}}             & {\ul 0.444$\pm$0}        \\
{SN-TS2Vec, cos}         & {\textbf{0.476$\pm$0}}     & {{\ul 0.306$\pm$0.142}}   & \textbf{0.663$\pm$0.089}  \\
{SN-TS2Vec, MMD}         & {\textbf{0.476$\pm$0}}     & {\textbf{0.467$\pm$0}}             & {\ul 0.444$\pm$0}        \\ \cline{1-4}
\end{tabular}
\label{table:ts2vec-vs-sn-ts2vec}
\end{table}

\subsubsection{TS-BYOL with and without spectral normalization}

\begin{figure}[h]
\centerline{\includegraphics[scale=0.29]{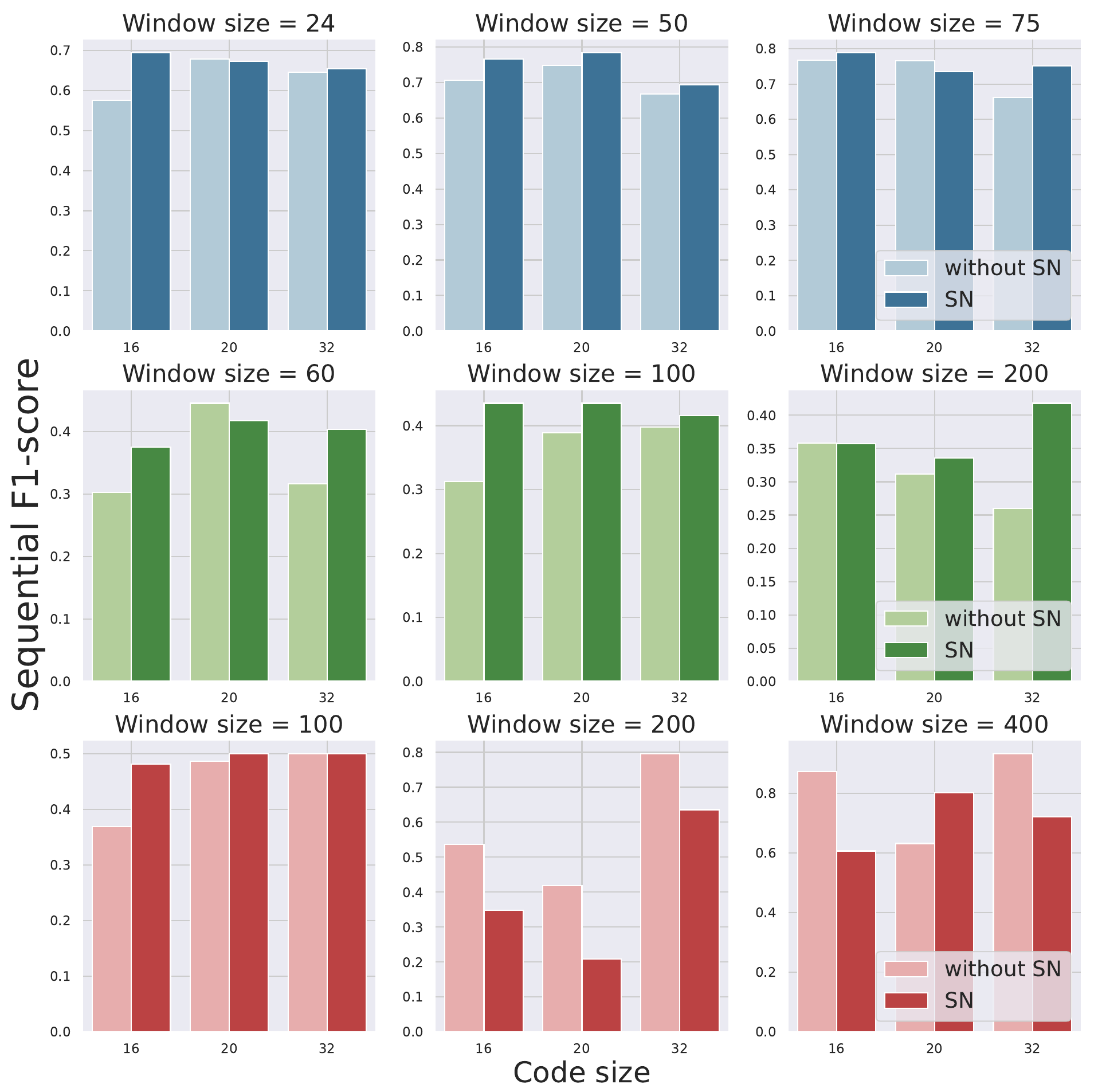}}
\caption{Comparison of TS-BYOL performance with and without spectral normalization on all datasets. The upper row~---~Yahoo!A4Benchmark, middle~---~HASC, lower~---~USC-HAD.}
\label{fig:byol_results}
\end{figure}

To enrich our comparison, we conducted additional experiments with an alternative self-supervised approach, TS-BYOL, on all three datasets. The basic TS-BYOL was compared to the TS-BYOL with SN. The results can be seen in Figure \ref{fig:byol_results}. It is clear that TS-BYOL with SN either outperforms or performs on par with its vanilla version on two of three datasets~---~Yahoo!A4Benchmark and HASC. The results obtained on the USC-HAD dataset are different; however, it should be noted that the total number of change points in the test set of USC-HAD is the smallest among all datasets, and the standard deviation of the obtained metrics is relatively high (up to $\text{std} = 0.4$). Therefore, the results on USC-HAD are not as representative as those of the other two datasets. 

\subsection{The dynamics of the representations}

\begin{figure}[htbp]\label{cos_dist}
\centering
\includegraphics[scale=0.25]{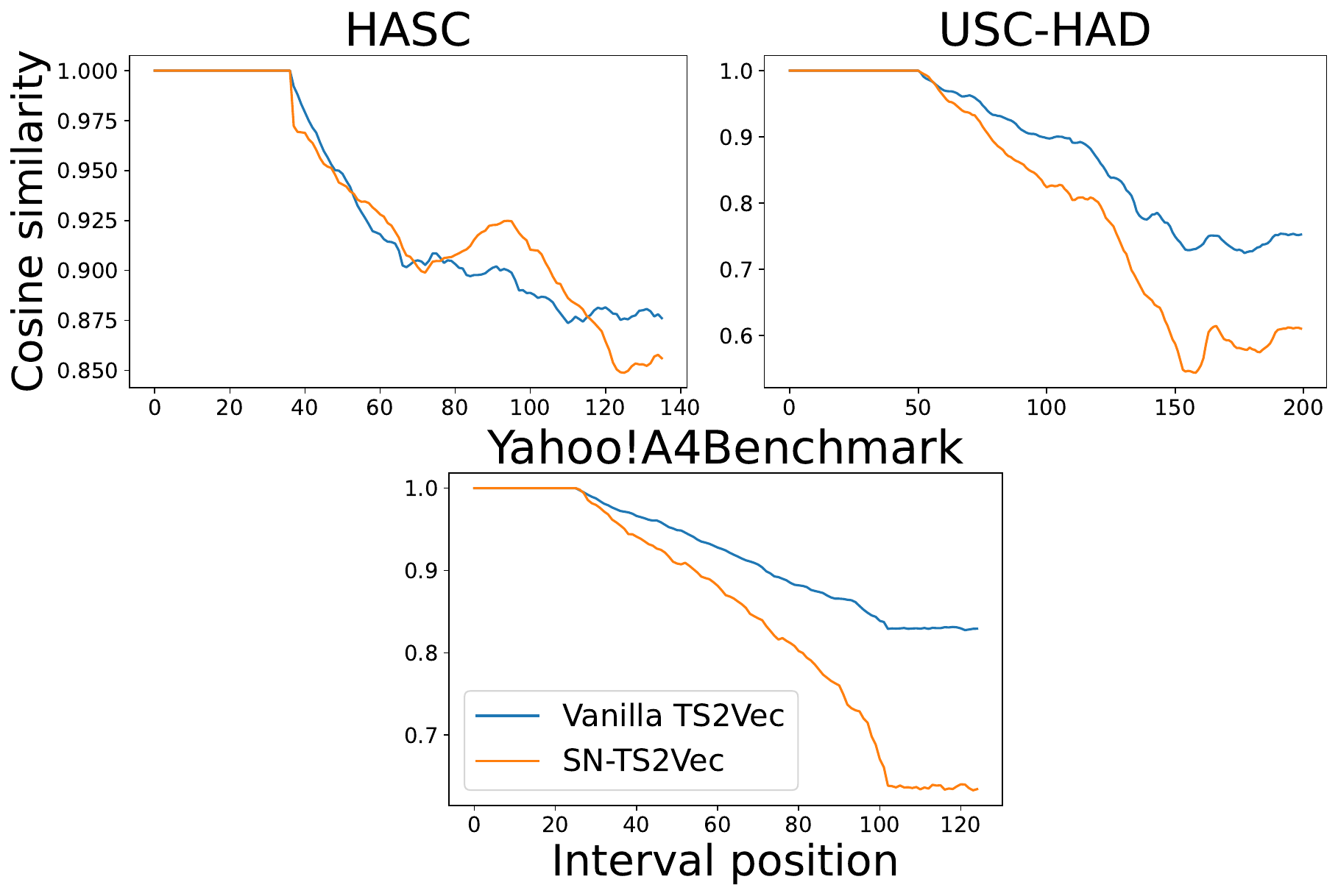}
\caption{The dynamics of the representations. Blue color corresponds to vanilla TS2Vec, and orange~---~to its SN version. Sliding window sizes: 60 for HASC, 75 for Yahoo!A4Benchmark, 100 for USC-HAD. Each value at the position $i$ corresponds to the similarity between the representations of $\mathbf{X}[i : i + w]$ and $\hat{\mathbf{X}}[i : i + w]$.}
\label{fig:dynamics}
\end{figure}

We compared the dynamics of the representations for Vanilla TS2Vec and SN-TS2Vec. The experiment was performed as follows:
\begin{enumerate}
    \item For each change point, we sampled a subsequence of length 300 right before the CP; let us call it $\mathbf{X}$. 
    \item We cloned each $\mathbf{X}$ and replaced the last half with the subsequence from right after the CP, denoting the sample obtained by $\hat{\mathbf{X}}$. Thus, for each CP, we have $\mathbf{X}$ and $\hat{\mathbf{X}}$: their first halves are identical, while their last halves belong to different distributions. 
    \item After that, we transformed $\mathbf{X}$ and $\hat{\mathbf{X}}$ into a sequence of representations via the sliding window procedure.
    \item  Finally, we calculated cosine similarities between the corresponding representations in $\mathbf{X}$ and $\hat{\mathbf{X}}$, obtaining a sequence of similarities for each CP. Supposedly, they should begin to diverge right after a CP appears.
\end{enumerate}
We averaged the obtained similarity arrays over all the change points. The results are presented in Figure \ref{fig:dynamics}. 
For two out of three considered datasets, the embeddings obtained with SN-TS2Vec diverge faster, achieving greater dissimilarity values. This visualization confirms that SN-TS2Vec is more suitable for CPD tasks than its vanilla version.

\section{Acknowledgments}

The research was supported by the Russian Science Foundation grant 20-7110135.

\section{Conclusion}
Our findings provide a step forward in the CPD field, bridging the gap between the expressive ability of representation learning models and the theoretical grounds of CPD problems.

We showed how to design self-supervised learning models with respect to CPD task specificity. 
The main contribution is the usage of Spectral Normalization, which, as we theoretically and empirically demonstrate, preserves test power for various two-sample tests in latent space. Thus, the resulting representations appear to be more robust for further change detection. 

Through a series of experiments, we confirm the effectiveness of the self-supervised approach, showing that the modern TS2Vec model outperforms current state-of-the-art approaches for CPD.  
Moreover, Spectral Normalization enhances the performance of TS2Vec, leading to a 5\% improvement in the target F1 metrics. 


\bibliographystyle{ieeetr}
\bibliography{bib}

\appendices
\section{Proofs from the section \ref{sec:methodology}}

\subsection{Proof of proposition \ref{prop:lr_preserve}}\label{app:proposition}
The proof is provided for the case of Gaussian random variables, but it is clear that the similar calculations may be performed for any distribution from the elliptical family.  

$\mathbf{X}$ can be viewed as a sample from the matrix normal distribution \cite{matrix_normal}, and the joint PDFs $p_0$ and $p_{\infty}$ have the densities of the following form (with the means $\mathbf{M}_0$ and $\mathbf{M}_{\infty}$, accordingly):
\begin{align}
    & p(\mathbf{X} | \mathbf{M}, \, \mathbf{U}, \, \mathbf{V}) = \\ & =  C \cdot \exp \Bigl(-\frac{1}{2} \text{tr} \left[\mathbf{V}^{-1}(\mathbf{X} - \mathbf{M})^T\mathbf{U}^{-1}(\mathbf{X}-\mathbf{M}) \right]\Bigr). \nonumber
\end{align}
First, consider $g(\mathbf{X}) = \mathbf{A}\mathbf{X} + \mathbf{b}$. It is a linear transformation, so the distribution of $g(\mathbf{X})$ can be written out explicitly:
\begin{align}
     &p(\mathbf{AX} | \mathbf{AM}, \mathbf{A}\mathbf{U}\mathbf{A}^T, \mathbf{V}) = \\ \nonumber &= C \cdot \exp \Bigl(-\frac{1}{2} \text{tr} \left[\mathbf{V}^{-1}(\mathbf{X} - \mathbf{M})^T \mathbf{A}^T \mathbf{A}^{-T}\mathbf{U}^{-1}\mathbf{A}^{-1}\mathbf{A} \cdot \right.  \\ \nonumber & \left. \cdot (\mathbf{X} - \mathbf{M}) \right]\Bigr) = \Tilde{C} \cdot p(\mathbf{X} | \mathbf{M}, \, \mathbf{U}, \, \mathbf{V}).
\end{align}
Here we let $b = 0$ to simplify the calculations. This implies that the linear transformation preserves the likelihood-ratio of the intervals. Now let us consider the invertible component $h(X)$. We know that for invertible transformations the PDF of the transformed distribution is calculated as
\begin{equation}
    \hat{p}(\mathbf{y}) = p(h^{-1}(\mathbf{y})) |J_{h^{-1}}(\mathbf{y})|,
\end{equation}
where $J_{h^{-1}}$ is the Jacobian matrix of the inverse transformation $h^{-1}$. Then the likelihood-ratio after applying $h$ becomes
\begin{align}
    \frac{\hat{p}_0(\mathbf{Y})}{\hat{p}_{\infty}(\mathbf{Y})} = \frac{p_0(\mathbf{X}) |J_{h^{-1}}(\mathbf{Y})|}{p_{\infty}(\mathbf{X}) |J_{h^{-1}}(\mathbf{Y})|} = \frac{p_0(\mathbf{X})}{p_{\infty}(\mathbf{X})},
\end{align}
i.e. it remains the same.

\subsection{Proof of lemma \ref{lem:test_power}}\label{app:lemma}
As was shown in Proposition \ref{prop:lr_preserve}, spectral normalized neural networks of a form \eqref{func} preserve the likelihood ratio, i.e., $\Lambda(\mathbf{X}) = \hat{\Lambda}(\mathbf{Y})$, where $\Lambda(\mathbf{X})$ and $\hat{\Lambda}(\mathbf{Y})$ denote the likelihood ratio in the raw data space and the representation space, accordingly. Therefore, it is true that for any function of the likelihood ratio $S(\Lambda)$ and a predefined threshold $h$
\begin{equation}\label{eq:equivalence}
    S(\Lambda(\mathbf{X})) > h \Longleftrightarrow S(\hat{\Lambda}(\mathbf{Y})) > h.
\end{equation}
The rejection of the null hypothesis is usually equivalent to the statistic value exceeding the threshold; hence, due to \eqref{eq:equivalence}, the rejection of the null hypothesis in the raw data space coincides with its rejection in the representation space.

\subsection{Proof of the RBF kernel distance preservation}\label{app:rbf}
The proof is given for $G: \mathbb{R}^{w \times D} \rightarrow \mathbb{R}^d$; the same calculations for a two-dimensional codomain of $G$ can also easily be conducted. Given:
\begin{align}
    L_1 \lVert & [\vecX_{i - w}, \dots, \vecX_i] -  [\vecX_{j - w}, \dots, \vecX_j] \rVert_2 \leq \lVert \vecY_i - \vecY_j \rVert_2 \leq \nonumber \\& \leq L_2 \lVert [\vecX_{i - w}, \dots, \vecX_i] -  [\vecX_{j - w}, \dots, \vecX_j] \rVert_2
\end{align}
Therefore, taking into account that the observations lie on the unit sphere and the $L_2$-norm properties, 
\begin{equation}
    L_1^2 \lVert \vecX_i - \vecX_j \rVert^2_2 \leq \lVert \vecY_i - \vecY_j \rVert^2_2 \leq L_2^2 (C + \lVert \vecX_i - \vecX_j \rVert^2_2),
\end{equation}
where $C > 0$ (a constant). 
Consequently, 
\begin{align}
\exp&{\Bigl(-\frac{\lVert \vecY_i - \vecY_j \rVert^2_2}{2\sigma^2}\Bigr)} \leq \exp{\Bigl(-\frac{L_1^2\lVert \vecX_i - \vecX_j \rVert^2_2}{2\sigma^2} \Bigr)} \leq \nonumber \\
&\leq \tilde C \exp{\Bigl(-\frac{\lVert \vecX_i - \vecX_j \rVert^2_2}{2\sigma^2} \Bigr)}.
\end{align}
where $\tilde C$ is some positive constant depending only on $L_1$ and $\sigma^2$. The last inequality uses the fact that the observations belong to the $\mathbb{S}^{D - 1}$. 
Similarly,
\begin{equation}
    \exp{\left(-\frac{\lVert \vecY_i - \vecY_j \rVert^2_2}{2\sigma^2}\right)} \geq \hat{C} \exp{\left(-\frac{\lVert \vecX_i - \vecX_j \rVert^2_2}{2\sigma^2}\right)}.
\end{equation}
Thus, the RBF kernel distance is indeed preserved.
\subsection{Proof of Theorem \ref{lem:mmd_lemma}}\label{app:mmd_lemma}
We suppose that the kernel $k$ is preserved under $G$, i.e. $\exists \, \bar{C} > 0, \tilde C > 0$ such that
\begin{equation}\label{eq:ineq}
    \hat{C}\cdot k(\vecX_i, \vecX_j) \leq k(\vecY_i, \vecY_j) \leq \tilde C \cdot k(\vecX_i, \vecX_j),
\end{equation}
where $\vecY_i = G([\vecX_{i - w}, \dots, \vecX_{i}])$. 

Denote $g_{w}(\vecX_i) = G([\vecX_{i - w}, \dots, \vecX_{i}])$. The kernel $k(\vecY_i, \vecY_j)$ in the space of the embeddings is equivalent to the kernel $\tilde{k}(\vecX_i, \vecX_j) = k(g_w(\vecX_i), g_w(\vecX_j))$ in the space of observations. Therefore, if $k(\vecX_i, \vecX_j) \leq K \; \forall \vecX_i, \vecX_j \in \mathcal{X}$, from \eqref{eq:ineq} immediately follows that the upper bound $\tilde{k}(\vecX_i, \vecX_j) \leq \tilde{C}K \; \forall \vecX_i, \vecX_j \in \mathcal{X}$ holds. 

Consider an $\mathrm{MMD}_b$-based hypothesis test of level $\alpha$ with the kernel $k$ s.t. $0 \leq k(x, y) \leq K$. It is shown in \cite{mmd} that this test has the acceptance region $$\mathrm{MMD}_b(Z, \Xi) < \sqrt{\frac{2K}{n}} \bigl(1 + \sqrt{2\log\alpha^{-1}} \bigr),$$ and is consistent:
\begin{align}\label{eq:test_consistency}
    \mathbb{P}\bigl\{|\mathrm{MMD}_b(Z, \Xi) & - \mathrm{MMD}_k(P, Q)| > 4\left(\frac{K}{n}\right)^{\frac{1}{2}} + \varepsilon\bigr\} \leq \\ \leq & 2 \exp{\left(\frac{-\varepsilon^2n}{4K}\right)}. \nonumber
\end{align}

The equation \eqref{eq:test_consistency} shows that the type II error converges to zero at rate $O(n^{-1/2})$; note that the change of constant $K$ to another constant $\tilde{C}K$ does not affect the convergence rate. Therefore, for the $\mathrm{MMD}_b$-based hypothesis test of level $\alpha$, the type II error convergence rate in the space of embeddings is the same as in the space of observations.

\section{Additional experiments}
\begin{figure}[htbp]
\centering
\includegraphics[scale=0.25]{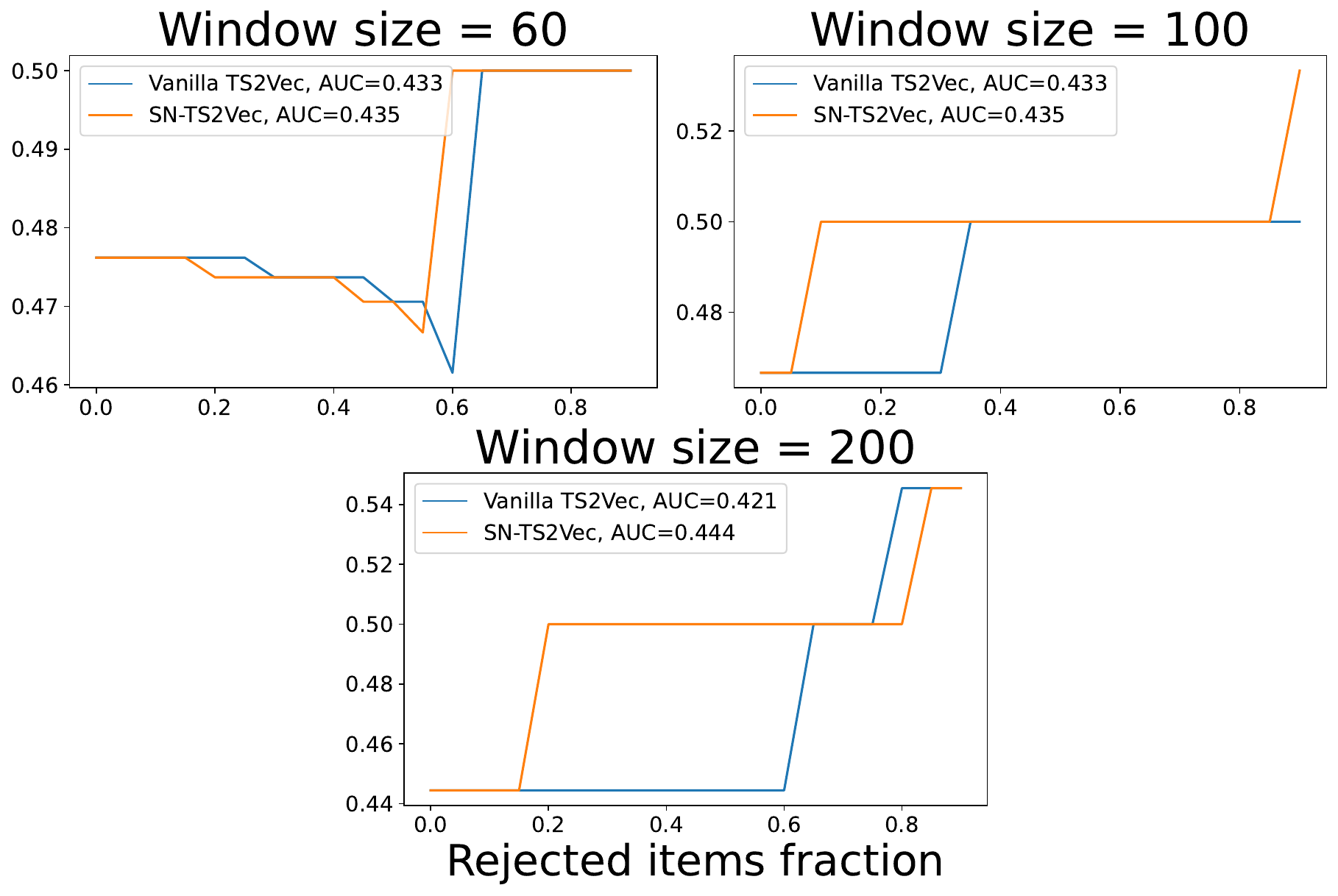}
\caption{F1-score rejection curves on HASC dataset. Blue color corresponds to vanilla TS2Vec, and orange — to its SN version. Higher AUCs correspond to better quality.}
\label{fig:rej_curve}
\end{figure}

An additional experiment investigates how the SN affects embedding properties from an uncertainty estimation perspective. We conducted the experiment as follows, adopting pipeline for a classification problem \cite{lee2018simple}:
\begin{itemize}
    \item Fit a multivariate Gaussian to the embeddings of train/val samples (windows tested for a presence of CP). 
    \item Calculate the Mahalanobis distance to this Gaussian for each test sample.
    \item Discard the top 5\% of test samples with the highest distances, removing the least "confident" samples, and measure the $F_1$-score for the remaining data. This process is repeated iteratively, reducing the dataset to 5\% of its original size.
\end{itemize}
Figure \ref{fig:rej_curve} shows the $F_1$-score rejection curves for a fixed seed, comparing SN-TS2Vec with its vanilla version on the HASC dataset. The SN version consistently outperforms the vanilla one, confirming its superiority in CPD tasks.

\section{Reproduction details}
\subsection{Train/val/test split}
\begin{table}[h]\label{tab:partition}
\caption{Train, val, and test set sizes}
\centering
\begin{tabular}{cccc}
\hline
\textbf{Dataset}  & \textbf{Train} & \textbf{Val} & \textbf{Test} \\ \hline
Yahoo!A4Benchmark & 0.4            & 0.2          & 0.4           \\
USC-HAD           & 0.6            & 0.2          & 0.2           \\
HASC              & 0.6            & 0.2          & 0.2           \\ \hline
\end{tabular}
\end{table}
Each dataset was split into three parts, according to the Table \ref{tab:partition}. The size of the test set was the same as in the TS-CP2 paper \cite{tscp} for the sake of the comparison. 

\subsection{Hyperparameter choice}
\subsubsection{TS-BYOL} A four-layer 1D convolutional backbone with ReLU and dropout feeds two-layer MLP projection and prediction heads. Trained for 10 epochs (validation every 5), the best validation checkpoint was used for testing; average over $3$ runs.
\subsubsection{SN-TS2Vec}  
TS2Vec model parameters: $depth=8$, $hidden\_dims=128$, $output\_dims=code\_size$. Training, vanilla: 40 epochs (Yahoo!), 10 epochs (HASC), $1$ epoch (USC-HAD). SN version: 40 epochs (Yahoo!), 5 epoch (HASC), 1 epoch (USC-HAD). Validation per epoch; the best validation checkpoint was used for testing. Averaged over 3 runs.
\subsection{Technical details}
All experiments were carried out on the NVidia L40.

\end{document}